\pdfoutput=1

\documentclass[11pt]{article}

\usepackage[final]{acl}

\usepackage{times}
\usepackage{latexsym}
\usepackage{amsmath}
\usepackage{mathtools}
\usepackage{bbm}
\usepackage{algorithm}
\usepackage{algorithmic}
\usepackage{amssymb}
\usepackage{multirow}
\usepackage{array}
\usepackage{float}
\usepackage{enumitem}
\usepackage{bbding}
\usepackage{makecell}
\usepackage{hyperref}

\usepackage{csquotes}

\usepackage[T1]{fontenc}

\usepackage[utf8]{inputenc}

\usepackage{microtype}

\usepackage{inconsolata}

\usepackage{graphicx}
\usepackage{ulem}

\title{Math-LLaVA: Bootstrapping Mathematical Reasoning for Multimodal Large Language Models}

\author{
  Wenhao Shi\textsuperscript{1}\thanks{Equal Contribution.}, Zhiqiang Hu\textsuperscript{2{$\ast$}}, Yi Bin\textsuperscript{3,4}\thanks{The corresponding author, email: yi.bin@hotmail.com}, Junhua Liu\textsuperscript{2}, Yang Yang\textsuperscript{1} \\
  \bf See-Kiong Ng\textsuperscript{4}, Lidong Bing, Roy Ka-Wei Lee\textsuperscript{2}\\
  \textsuperscript{1}University of Electronic Science and Technology of China \\
  \textsuperscript{2}Singapore University of Technology and Design \\
\textsuperscript{3}Tongji University \textsuperscript{4}National University of Singapore\\
}

\begin{document}
\maketitle
\begin{abstract}

Large language models (LLMs) have demonstrated impressive reasoning capabilities, particularly in textual mathematical problem-solving. However, existing open-source image instruction fine-tuning datasets, containing limited question-answer pairs per image, do not fully exploit visual information to enhance the multimodal mathematical reasoning capabilities of Multimodal LLMs (MLLMs). To bridge this gap, we address the lack of high-quality, diverse multimodal mathematical datasets by collecting 40K high-quality images with question-answer pairs from 24 existing datasets and synthesizing 320K new pairs, creating the MathV360K dataset, which enhances both the breadth and depth of multimodal mathematical questions. We introduce Math-LLaVA, a LLaVA-1.5-based model fine-tuned with MathV360K. This novel approach significantly improves the multimodal mathematical reasoning capabilities of LLaVA-1.5, achieving a 19-point increase and comparable performance to GPT-4V on MathVista's minitest split,
and yielding leading performance on Math-V and MathVerse.
Furthermore, Math-LLaVA demonstrates enhanced generalizability, showing substantial improvements on the MMMU benchmark. Our research highlights the importance of dataset diversity and synthesis in advancing MLLMs' mathematical reasoning abilities.
The code and data are available at: \url{https://github.com/HZQ950419/Math-LLaVA}.

\end{abstract}

\section{Introduction}

\textbf{Motivation.} Large language models (LLMs) exhibit impressive reasoning capabilities, drawing significant research interest in mathematical problem-solving in textual form \cite{zhang2020graph, jason2022cotreason, wang2023selfconsistency, Bin2023mtreemath, luo2023wizard, yue2023mammoth, gou2023tora, zhou2023least-to-most}. However, the task of multimodal mathematical reasoning \cite{lu2023mathvista} requires models to interpret diverse images and apply advanced reasoning skills. While open-source multimodal large language models (MLLMs) like LLaVA \cite{liu2023llava} and Mini-GPT4 \cite{zhu2023minigpt4} perform well on VQA tasks \cite{guo2023unkvqa}, they fall short of proprietary MLLMs \citep{gpt4v, gemini} in solving complex mathematical problems involving visual content.

Two common approaches to enhance MLLMs' mathematical reasoning skills are prompt methods and fine-tuning methods. Prompt methods \cite{lu2023mathvista, wang2024all} leverage MLLMs' latent abilities through carefully designed prompts, while fine-tuning methods \cite{wang2024tsciq, hu2023visual, zheng2023ddcot} adjust model parameters using reasoning data collected from real-world or synthetic data from advanced LLMs (e.g., GPT-4). However, existing open-source image instruction fine-tuning datasets \cite{lu2022dynamic, li2023super, lu2022scienceqa}, which contain limited question-answer pairs per image, do not fully exploit visual information to enhance MLLMs' multimodal mathematical reasoning capabilities.

\textbf{Research Objectives.} To bridge this gap, we select 40K high-quality images with corresponding question-answer pairs from 24 pre-existing datasets. These images and queries span various subjects, including algebra, arithmetic, geometry, logic, numeric commonsense, science, and visual question answering. The selection criteria were based on image clarity and comprehension complexity. Additionally, we propose a pipeline to synthesize 320K new pairs based on the 40K images and seed inquiries.

Constructing such a dataset presents significant challenges, including selecting diverse and high-quality multimodal question-answer data and enhancing question diversity. Selecting suitable data involves filtering for image clarity and comprehension complexity, ensuring the dataset covers a wide range of mathematical concepts and question types. Enhancing question diversity requires synthesizing new questions that probe different aspects of the images and involve multiple reasoning steps. To further improve model robustness and comprehension, we focus on enhancing logical consistency \cite{tascon2023logical} and the ability to understand underspecified language \cite{pezzelle2023underspecified}.


\textbf{Contributions.} Using the selected 40K data, the fine-tuned LLaVA-1.5 model, named Math-LLaVA-DS, achieved a significant improvement of 10.6\% on MathVista \cite{lu2023mathvista}. To further enhance multimodal mathematical reasoning capabilities, we synthesized an additional 320K question-answer pairs based on the 40K images and seed questions, resulting in the MathV360K dataset. This comprehensive dataset, containing around 40K images and 360K question-answer pairs, significantly expands the coverage of multimodal mathematical reasoning. Fine-tuning LLaVA-1.5 with MathV360K, we developed Math-LLaVA, which outperforms the original LLaVA-1.5 by 19\% on MathVista's minitest split
and achieves leading performance on Math-V \cite{wang2024mathv} and MathVerse \cite{zhang2024mathverse}.
We also evaluated Math-LLaVA on MMMU \cite{yue2023mmmu}, demonstrating its improved generalizability. 
\begin{figure*}[t]
    \centering
    \includegraphics[width=0.95\textwidth]{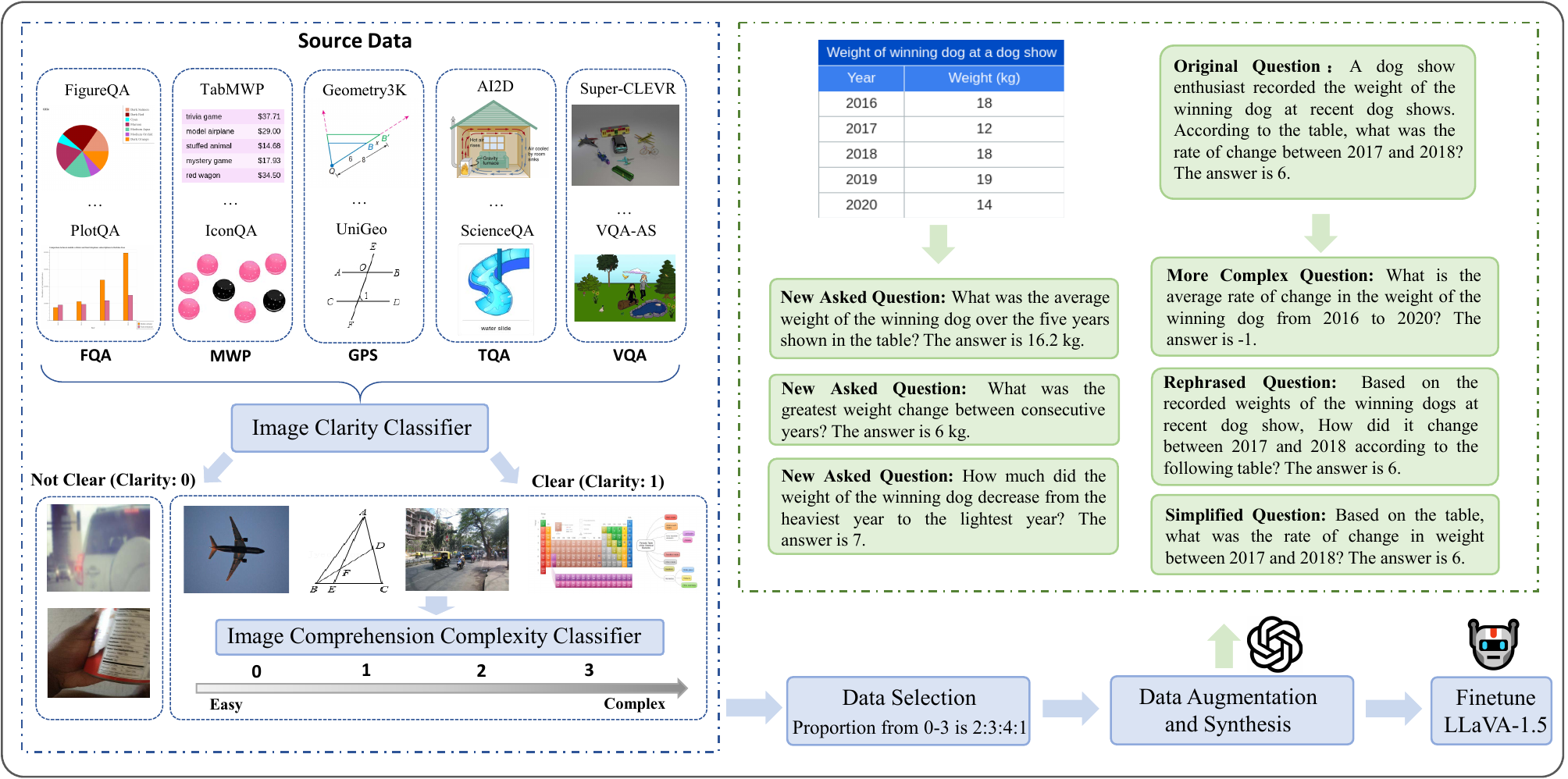}
    \caption{The overall flowchart of the proposed multimodal question-answer data selection and data augmentation. Our data selection depends on the fine-tuned ViT as image classifier. The data generation process depends on the vision-language model.
    }
    \vspace{-10pt}
    \label{fig:framework}
\end{figure*}

\section{Related Works}
\subsection{Multimodal Large Language Models}

The advancement of LLMs has spurred significant research interest in vision-language interaction, particularly in integrating visual knowledge into LLMs. The CLIP series \cite{radford2021clip, li2022blip} aligned visual and language modalities using contrastive learning on extensive image-text pairs. Recent studies increasingly use pre-training alignment and visual instruction tuning on LLMs for complex tasks like visual question answering, artwork analysis, and multimodal reasoning~\cite{li2024mm,bin2024gallerygpt}. MiniGPT-4 \cite{zhu2023minigpt4} engages in image-text dialogues by aligning visual features with text. Similarly, models like LLaVA \cite{liu2023llava} and InstructBLIP \cite{dai2024instructblip} use learnable projectors or query embeddings to interact with visual features. These approaches aimed to leverage high-quality pre-training and fine-tuning data to comprehend complex instructions. Models like mPLUG-Owl \cite{ye2023mplug}, SPHINX \cite{lin2023sphinx}, and MiniCPM-V2 \cite{hu2024minicpm} introduced new grounding data types and modularization training to minimize hallucinations and enhance grounding abilities. Despite these advancements, MLLMs face challenges in solving multimodal mathematical problems using diagrams. Further exploration of the quality and format of image instructions is needed to improve the reasoning capabilities of MLLMs.

\subsection{Multimodal Reasoning}
The rapid development of MLLMs has advanced research on multimodal reasoning \cite{chen2024plugplay, you2023ideagpt}. Augmenting the original question and answer text data in restricted domains to further fine-tune MLLMs is a popular approach. For raw answers, rationales were either generated by humans \cite{zhang2023multicot} or gathered from prominent LLMs \cite{wang2024tsciq, lin-etal-2023-beneath, chen2023chain, li2024-multimodal-arxiv}. Additionally, VPD \cite{hu2023vpd} proposed expanding answers by converting programming code formats to natural language formats. For raw questions, DDCoT \cite{zheng2023ddcot} used LLMs to decompose the original questions into sub-questions. These methods, however, only utilize LLMs to target text-only data within restricted domains, neglecting to fully exploit the visual information in raw images for further enhancement. To evaluate the multimodal reasoning abilities of MLLMs more comprehensively, MathVista \cite{lu2023mathvista}, 
Math-V \cite{wang2024mathv} and MathVerse \cite{zhang2024mathverse},
which involve various types of mathematical reasoning and skills, and MMMU \cite{yue2023mmmu}, which encompasses multidisciplinary tasks, have been proposed. There is still significant room for improvement in open-source MLLMs.

\section{Data Synthesis}



Existing open-source image instruction fine-tuning datasets \cite{lu2022dynamic, li2023super, lu2022scienceqa}, containing limited question-answer pairs per image, do not fully exploit visual information to enhance the multimodal mathematical reasoning capabilities of MLLMs. To address this, we propose MathV360K, a robust dataset synthesized based on the 40K selected images and seed question-answer pairs from multiple sub-domains. As shown in the left side of Figure~\ref{fig:framework}, we first select 40K high-quality data points based on the image clarity and comprehension complexity from 24 open-source multimodal question-answering datasets. In the second step, illustrated in the top right of Figure~\ref{fig:framework}, we attempt to fully mine the visual information of the images to generate additional questions. The data generation pipeline includes creating diverse new questions to fully exploit the visual information, more complex questions to further improve the reasoning capabilities, rephrased questions and underspecified questions to improve the robustness of the model. With the data generation pipeline, we collected 360K high-quality and diverse instruction-tuning data for the selected 40K data points to enhance the image understanding and mathematical reasoning capabilities of the LLaVA-1.5 open-source model.

\subsection{Multimodal Reasoning Data Selection} \label{Data Selection}
\subsubsection{Source Data}

We collected 24 visual question answering and multimodal mathematical reasoning datasets, each targeting a specific task type and visual content. We focused on five problem task types requiring high-level reasoning to compile the source dataset: Figure Question Answering (FQA), Geometry Problem Solving (GPS), Math Word Problem (MWP), Textbook Question Answering (TQA), and Visual Question Answering (VQA).
Table~\ref{tab:sourcedata} in Appendix shows more details about the task type and visual context of each source dataset.


Each multimodal training sample consists of three components: an image $I_i$, a text question $Q_i$, and a ground-truth answer $A_i$. From this data format, the model aims to capture visual information and question semantics to reason the final answer.

\subsubsection{Image Filtering and Proportioning} \label{Image Filtering and Proportioning}

After acquiring the 24 source datasets, we intentionally selected data from the raw images based on the following criteria: (1) The clarity of the images, as poor-quality images introduce noise and interfere with learning image semantics; (2) The comprehension complexity of the images, which varies from easy to complex. By categorizing images into different levels of complexity and selecting proportionally, we can form a training set with an appropriate difficulty distribution; (3) The quality of the corresponding textual question data, ensuring that the difficulty aligns with the comprehension complexity of the images.

\begin{figure}[h]
  \centering
\includegraphics[width=1.0\columnwidth]{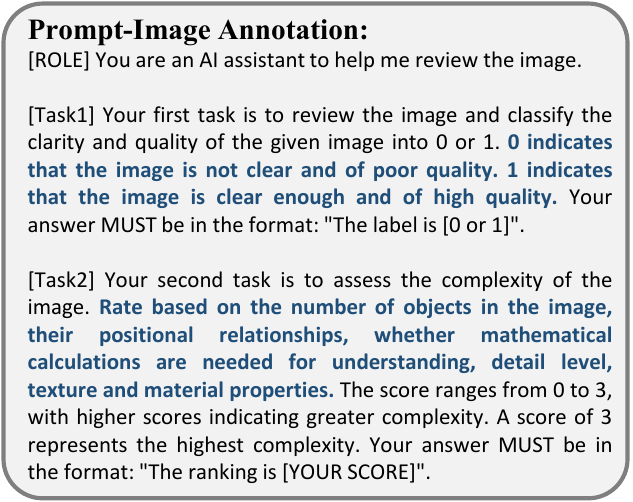}
  \caption{The prompt template used in our GPT-4V API for image annotation. Image clarity is considered as binary classification and image comprehension complexity is viewd as multi-classification.}
  \label{fig:prompt-annotate}
\end{figure}

We fine-tuned two Vision Transformer (ViT) \cite{dosovitskiy2021vit} models to classify image clarity and image comprehension complexity, respectively. Due to the lack of annotated image data, we first sampled 10K images uniformly and randomly from the source datasets. These images were labeled for clarity and comprehension complexity using GPT-4V \cite{gpt4v}, with our designed prompt shown in Figure~\ref{fig:prompt-annotate}. For image clarity, label 0 indicates a blurred, poor-quality image, and label 1 indicates a clear, good-quality image. Image comprehension complexity is determined by the number of objects, their positional relationships, the need for mathematical calculations, detail level, texture, and material properties. Images are categorized into scores of 0, 1, 2, and 3, with lower values indicating easier visual context comprehension.

Based on the 10K annotated images, we trained two ViT models with initialized fully connected layers for classification using cross-entropy loss. We first classified all source training dataset images using the fine-tuned image clarity classifier and filtered out images labeled as 0. Table~\ref{tab:sourcedata} shows the number of images before (i.e., \textit{Training Images}) and after (i.e., \textit{Clear Images}) filtering.

Next, we used the image comprehension complexity classifier to score the filtered images. Table~\ref{tab:sourcedata} shows that most images are classified as medium complexity, followed by easy, and finally the most complex. Considering that simple images are easier to learn from, while complex images are harder and require more reference samples, we sampled the first three complexity categories using a progressive scale from simple to complex. Since images with a score of 3 are the least abundant, we collected all of them. We selected 40K data points based on an overall ratio of complexity 2:3:4:1, ensuring samples from different complexities are uniformly selected from each source dataset. As a result, we obtained 40K high-quality ($I$, $Q$, $A$) real data points that are diverse in image information and questions are progressive in difficulty.


\begin{figure*}[]
    \centering
    \includegraphics[width=0.95\textwidth]{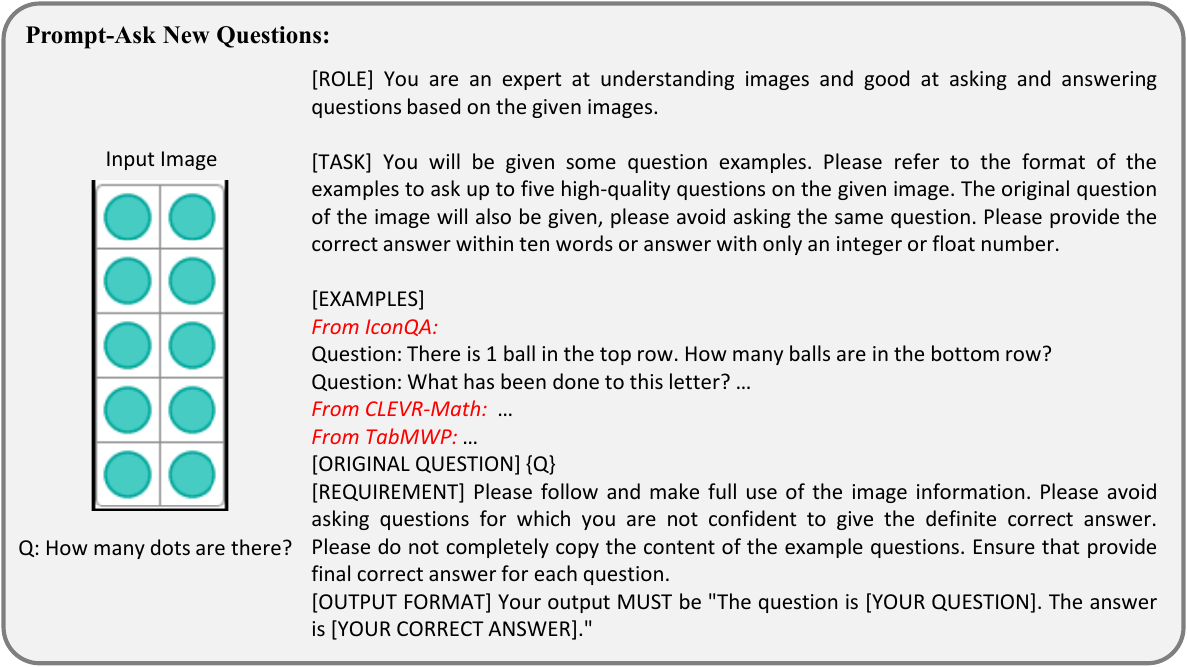}
    \caption{The prompt template used in our GPT-4V API generates additional questions for each input image. Demonstrations are constructed by randomly sampling one question from each cluster of each source dataset belonging to a specific task type. 
    }
    \label{fig:prompt-gene}
\end{figure*}

\subsection{Data Augmentation}
\subsubsection{Mining Image for QA Generation}

After selecting 40K multimodal reasoning data, we observed that each image typically corresponds to limited questions. As shown in the tabular image of Figure~\ref{fig:framework}, the original question often focuses only on localized arithmetic differences. However, additional questions about overall averages, continuous variations, and more can also be asked, indicating that the visual information of an image is not fully exploited with just one question. Therefore, we can further augment the available real data by generating more question-answer pairs for each image.


We use GPT-4V to generate additional questions based on the input image and the original question. If questions are generated in a zero-shot manner, they often focus on one-sided visual scenes, lacking reasoning and mathematical skills. For images from specific tasks, such as geometric figures, more task-specific questions should be asked. Therefore, we adopt few-shot demonstrations for GPT-4V to generate new questions.

\begin{figure}[h]
  \centering
\includegraphics[width=1.0\columnwidth]{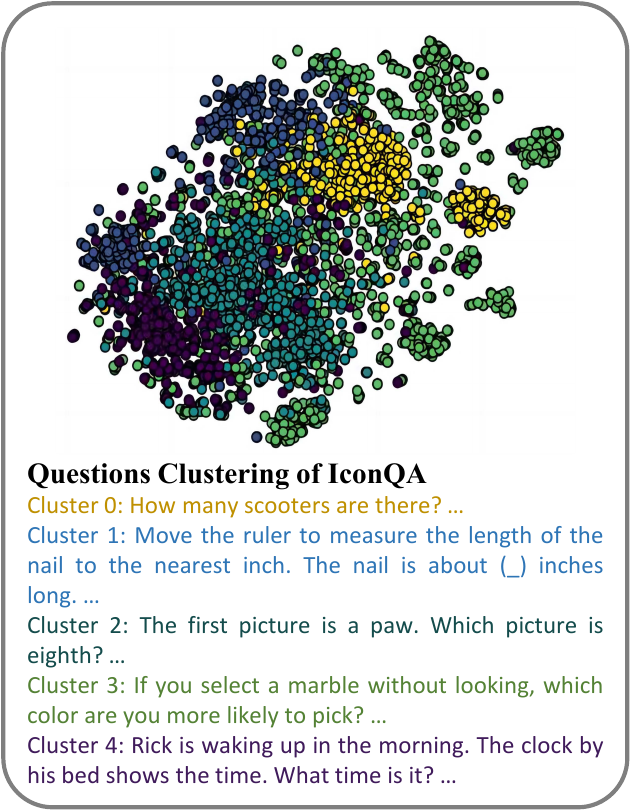}
  \caption{The visualization of the K-Means by T-SNE. We take IconQA as example. The questioning format of each cluster can be used as a reference to generate new questions for similar visual content.}
  \vspace{-10pt}
  \label{fig:cluster}
\end{figure}

For an image belonging to one of the categories (FQA, GPS, MWP, TQA, VQA), we first internally cluster the questions into five classes for each source dataset within that task category. Specifically, features of text questions are obtained using TF-IDF and clustered using K-Means. As shown in Figure~\ref{fig:cluster}, we take IconQA as an example. After clustering the questions in the training set, each cluster internally represents a specific questioning format and pattern that can be referenced. Demonstrations are constructed by randomly sampling one question from each cluster of each source dataset belonging to a certain task type.

The prompt for generating new questions for an input image is shown in Figure~\ref{fig:prompt-gene}. This method ensures that the newly generated questions are consistent with the distribution of the original reference questions while improving diversity. Using this approach, we generated 200K new question-answer pairs based on the selected 40K data points.

\subsubsection{Augmentation of Original Question}

\begin{figure}[]
    \centering
   \includegraphics[width=1.0\columnwidth]{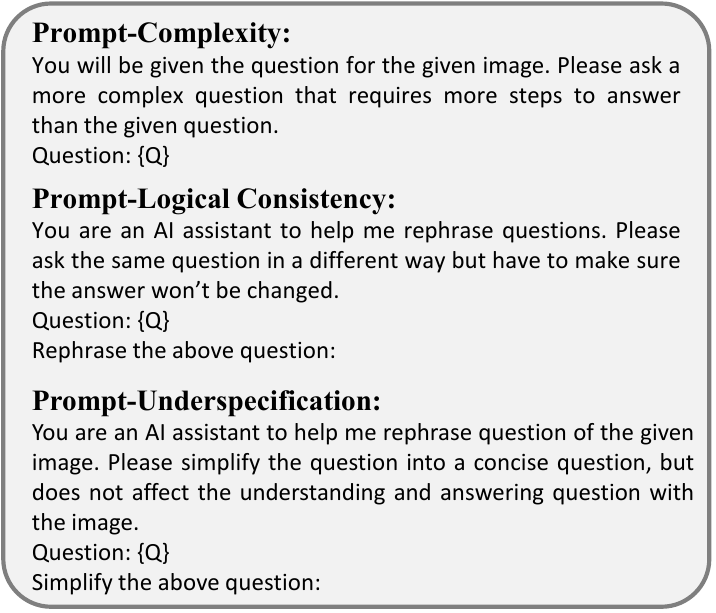}
    \caption{
    The prompt templates used in our GPT-4V API to generate more complex, logically consistent and underspecified questions from original question text. 
    }
    \label{fig:prompt-question}
\end{figure}

We designed prompts to augment the original questions, as shown in Figure~\ref{fig:prompt-question}. Using GPT-4V, we generated 40K more complex questions, 40K simplified questions, and 40K rephrased questions. The augmentation focused on the following aspects:

\noindent \textbf{Complexity}. More complex reasoning samples can enhance the reasoning capabilities of fine-tuned LLMs \cite{luo2023wizard}. Our first approach involves creating more complex questions based on the original image and corresponding inquiries.

\noindent \textbf{Logical Consistency}. Robust MLLMs should answer consistently about similar content in a given image \cite{tascon2023logical}. We employed GPT-4V to ask the same question in different ways without changing the answer.

\noindent \textbf{Underspecification}. Robust MLLMs must deal with semantic underspecification, where the linguistic signal conveys only part of the necessary information for successful communication \cite{pezzelle2023underspecified}. Therefore, we simplified the original questions without affecting their semantic understanding when combined with the image.

\section{Experiments}
\subsection{Model and Training}

We employ the LLaVA-1.5 architecture as our base model, which primarily comprises the Vicuna-v1.5 language model \cite{vicuna} and a pretrained Vision Transformer (ViT) as the image encoder. To preserve the foundational model's superior visual perception and descriptive abilities, we fine-tune LLaVA-1.5-13B using the proposed MathV360K instruction-tuning dataset. The diverse question patterns and rich visual content within this dataset enhance the model's multimodal mathematical reasoning capabilities while maintaining its general vision-language understanding skills.


\subsection{Evaluation and Metrics}
We evaluate our model using the minitest subset of MathVista \cite{lu2023mathvista} in a zero-shot manner. This minitest subset comprises 1,000 samples, including 540 multiple-choice questions and 460 questions that require free-form answers in the form of integers, floats, or lists. MathVista adequately assesses the MLLMs' multimodal mathematical skills, including algebraic reasoning (ALG), arithmetic reasoning (ARI), geometry reasoning (GEO), logical reasoning (LOG), numeric commonsense (NUM), scientific reasoning (SCI), and statistical reasoning (STA). Furthermore, MathVista questions can be categorized into the following subsets: FQA, GPS, MWP, TQA, and VQA. For evaluation, we first employ GPT-4 to extract the predicted choices or answers from responses, then report the answer accuracy, which entails determining whether the final answer matches the ground truth.
We also conduct evaluation using the
Math-V \cite{wang2024mathv} and MathVerse \cite{zhang2024mathverse}. Math-V is a meticulously curated
collection of 3,040 mathematical problems with visual contexts sourced from real math competitions. MathVerse consists of 2,612 multi-subject math problems varying degrees of information
content in multi-modality.
Additionally, we evaluate our model's enhanced generalizability using the MMMU benchmark \cite{yue2023mmmu}. The MMMU benchmark, with 900 evaluation samples, encompasses six core disciplines: Art \& Design, Business, Science, Health \& Medicine, Humanities \& Social Science, and Technology \& Engineering, making it suitable for assessing the generalization of MLLMs' reasoning capabilities.

\subsection{Implementation Details}
We utilize GPT-4V (GPT-4 Vision Preview) for the data generation process. To classify image clarity and comprehension complexity, we fine-tune two ViT-Large-Patch16-224 models, each with a learning rate of 2e-4 and a training period of 5 epochs. For the LLaVA-1.5-13B model, the input image resolution is configured to 336 by 336 pixels. Both the projection linear layer and the language model are trainable. During the fine-tuning phase, we set a learning rate of 2e-5, employ a batch size of 16, and conduct fine-tuning over 2 epochs using A800 GPUs equipped with 80GB of memory.

\section{Results and Analysis}
\subsection{Main Comparison on MathVista}
We compare Math-LLaVA with other MLLMs on the minitest split of the MathVista benchmark in Table~\ref{tab:baseline}. As shown in the table, open-source MLLMs such as miniGPT4 \cite{zhu2023minigpt4}, instructBLIP \cite{dai2024instructblip}, and LLaVA-1.5-13B have poor performance in multimodal mathematics, with overall accuracy lower than 30\%. Compared to the base model, LLaVA-1.5-13B, with poor multimodal mathematical ability, Math-LLaVA achieves 46.6\% overall accuracy with a significant improvement of 19\%. More surprisingly, the proposed Math-LLaVA model outperforms close-source models Gemini 1.0 Pro \cite{team2023gemini} and Claude 3 Haiku \cite{anthropic2024claude}, even achieving comparable performance to GPT-4V \cite{gpt4v}, the most powerful close-source MLLMs.
Interestingly, Math-LLaVA achieves 57.7\% accuracy on GPS subset, outperforming G-LLaVA-13B \cite{gao2023gllava}, which has been trained on 170K high-quality geometric image-caption and question-answer pairs. 
The results on Math-V are shown in Table~\ref{tab:math-v}.
Math-LLaVA also achieves significant improvement compared to the base model and leading performance than Qwen-VL-Max \cite{bai2023qwenvl} and most open-source MLLMs.
The results on MathVerse could be found from Table~\ref{tab:mathverse} in Appendix.
The superior performance of Math-LLaVA indicates that the data selection and synthesis of high-quality, diverse multimodal question-answer pairs are effective in improving MLLM's multimodal mathematical reasoning capabilities.

\begin{table*}[t]\small
\centering

\renewcommand{\arraystretch}{1.3}
\setlength{\tabcolsep}{1.0mm}{
\begin{tabular}{c|c|ccccc|ccccccc}
\hline

\multicolumn{1}{c|}{\multirow{2}{*}{\textbf{Model}}} & \multicolumn{13}{c}{\textbf{MathVista}} \\ \cline{2-14}
\multicolumn{1}{c|}{} & ALL& FQA& GPS& MWP& TQA& VQA& ALG &ARI &GEO &LOG &NUM &SCI &STA         \\ \hline

\multicolumn{14}{c}{\multirow{1}{*}{\textit{Heuristics Baselines}}} \\ \hline
\multicolumn{1}{c|}{Random Chance} & 17.9& 18.2& 21.6& 3.8& 19.6& 26.3& 21.7 &14.7 &20.1 &13.5 &8.3 &17.2 &16.3         \\
\multicolumn{1}{c|}{Frequent Guess \cite{lu2023mathvista}} & 26.3& 22.7& 34.1& 20.4& 31.0& 24.6& 33.1 &18.7 &31.4 &24.3 &19.4 &32.0 &20.9         \\
\multicolumn{1}{c|}{Human} & 60.3&	59.7&	48.4&	73.0&	63.2&	55.9&	50.9	&59.2&	51.4&	40.7&	53.8&	64.9&	63.9         \\ \hline

\multicolumn{14}{c}{\multirow{1}{*}{\textit{Close-Source Multimodal Large Langugae Models (MLLMs)}}} \\ \hline
\multicolumn{1}{c|}{Gemini 1.0 Nano 2 \cite{team2023gemini}} & 30.6&	28.6&	23.6&	30.6&	41.8&	31.8&	27.1&	29.8&	26.8&	10.8&	20.8&	40.2&	33.5      \\
\multicolumn{1}{c|}{Qwen-VL-Plus \cite{bai2023qwenvl}} &43.3 &	\textbf{54.6} &	38.5&	31.2&	55.1 & 34.1 & 39.1&	32.0&	39.3&	18.9 &	26.4&	59.0 &	56.1\\ 
\multicolumn{1}{c|}{Gemini 1.0 Pro \cite{team2023gemini}} & 45.2&	47.6&	40.4&	39.2&	61.4&	\textbf{39.1}&	45.2&	38.8&	41.0&	10.8&	\textbf{32.6}&	54.9&	\textbf{56.8}     \\
\multicolumn{1}{c|}{Claude 3 Haiku \cite{anthropic2024claude}} & 46.4&	-&	-&	-&	-&	-&	-&	-&	-&	-&	-&	-&	-     \\
\multicolumn{1}{c|}{GPT-4V \cite{gpt4v}} & \textbf{49.9}&	43.1&	\textbf{50.5}&	\textbf{57.5}&	\textbf{65.2}&	38.0&	\textbf{53.0}&	\textbf{49.0}&	\textbf{51.0}&	\textbf{21.6}&	20.1&	\textbf{63.1}&	55.8    \\
 \hline

\multicolumn{14}{c}{\multirow{1}{*}{\textit{Open-Source Multimodal Large Langugae Models (MLLMs)}}} \\ \hline
\multicolumn{1}{c|}{mPLUG-Owl-7B \cite{ye2023mplug}} & 22.2& 22.7& 23.6& 10.2& 27.2& 27.9& 23.6 &19.2 &23.9 &13.5 &12.7 &26.3 &21.4 \\
\multicolumn{1}{c|}{miniGPT4-7B \cite{zhu2023minigpt4}} & 23.1&	18.6&	26.0& 13.4&	30.4&	30.2&	28.1&	21.0&	24.7&	16.2&	16.7&	25.4&	17.9 \\
\multicolumn{1}{c|}{LLaVAR-13B \cite{zhang2023llavar}} & 25.2&	21.9&	25.0&	16.7&	34.8&	30.7&	24.2&	22.1&	23.0&	13.5&	15.3&	42.6&	21.9 \\
\multicolumn{1}{c|}{InstructBLIP-7B \cite{dai2024instructblip}} & 25.3&	23.1&	20.7&	18.3&	32.3&	35.2&	21.8&	27.1&	20.7&	18.9&	20.4&	33.0&	23.1 \\
\multicolumn{1}{c|}{LLaVA-13B \cite{liu2023llava}} & 26.1&	26.8&	29.3&	16.1&	32.3&	26.3&	27.3&	20.1&	28.8&	24.3&	18.3&	37.3&	25.1 \\
\multicolumn{1}{c|}{SPHINX-V1-13B \cite{lin2023sphinx}} & 27.5&	23.4&	23.1&	21.5&	39.9&	34.1&	25.6&	28.1&	23.4&	16.2&	17.4&	40.2&	23.6 \\
\multicolumn{1}{c|}{LLaVA-1.5-13B \cite{liu2024llava-1.5}} & 27.6&	-&	-&	-&	-&	-&	-&	-&	-&	-&	-&	-&	- \\
\multicolumn{1}{c|}{LLaVA-1.5-13B$^\dagger$ \cite{liu2024llava-1.5}} & 27.7&	23.8&	22.7&	18.3&	40.5&	30.2&	25.3&	26.4&	22.8&	21.6&	26.4&	35.3&	23.6 \\
\multicolumn{1}{c|}{OmniLMM-12B \cite{omnilmm}} & 34.9&	45.0&	17.8&	26.9&	44.9&	39.1&	23.1&	32.3&	20.9&	18.9&	27.8&	45.9&	44.2 \\
\multicolumn{1}{c|}{SPHINX-V2-13B \cite{lin2023sphinx}} &36.7&	\textbf{54.6} &	16.4&	23.1&	41.8&	\textbf{43.0}&	20.6&	33.4&	17.6&	\textbf{24.3}&	21.5&	43.4&	\textbf{51.5}\\

\multicolumn{1}{c|}{G-LLaVA-13B \cite{gao2023gllava}} &- &	- &	56.7&	-&	-& - & -&	-&	-&	- &	-&	- &	-\\ 
\hline

\multicolumn{1}{c|}{Math-LLaVA-DS} &38.2& 	33.5&	47.2&	41.4&	36.7&	34.6&	38.4&	34.3&	45.6&	18.9&	33.3&	45.9&	35.2 \\
\multicolumn{1}{c|}{\textbf{Math-LLaVA}} & \textbf{46.6}	& 37.2	& \textbf{57.7} &	\textbf{56.5}	& \textbf{51.3}	&33.5 &	\textbf{53}	& \textbf{40.2} &	\textbf{56.5} &	16.2 &	\textbf{33.3} &	\textbf{49.2} &	43.9 \\ \hline

\end{tabular}}
\caption{Comparison with baselines on the testmini set of MathVista benchmark. Baseline results are
obtained from \citet{lu2023mathvista}. 
$^\dagger$ represents our reproduced results of LLaVA-1.5-13B.
The best results in both the close-source and open-source MLLMs are in bold. MathVista is divided in two ways: task type or mathematical skill, and we report the accuracy under each subset.}
\vspace{-10pt}
\label{tab:baseline}
\end{table*}

\begin{table*}[h]\small
\centering

\renewcommand{\arraystretch}{1.4}
\setlength{\tabcolsep}{0.30mm}{
\begin{tabular}{c|c|cccccccccccccccc}
\hline

\multicolumn{1}{c|}{\multirow{2}{*}{\textbf{Model}}} & \multicolumn{16}{c}{\textbf{Math-V}} \\ \cline{2-18}
\multicolumn{1}{c|}{} &ALL &Alg &AnaG& Ari& CG& Comb& Cnt& DG& GT& Log& Angle& Area& Len& SG &Sta& Topo& TG         \\ \hline
\multicolumn{18}{c}{\multirow{1}{*}{\textit{Heuristics Baselines}}} \\ \hline
\multicolumn{1}{c|}{Random Chance} &7.2& 1.5& 11.9& 7.1& 9.7 &4.8& 6.0& 22.1& 1.1& 7.6& 0.6& 9.4& 6.7& 8.2 &8.6& 13.0& 7.1    \\
\multicolumn{1}{c|}{Human} & 68.8 &	55.1&	78.6&	99.6&	98.4&	43.5&	98.5&	91.3&	62.2&	61.3&	33.5&	47.2&	73.5&	87.3&	93.1&	99.8&	69.0         \\ \hline

\multicolumn{18}{c}{\multirow{1}{*}{\textit{Close-Source Multimodal Large Langugae Models (MLLMs)}}} \\ \hline

\multicolumn{1}{c|}{Qwen-VL-Plus \cite{bai2023qwenvl}} &10.7&	11.3&	17.9&	14.3&	12.7&	4.8&	10.5&	15.4&	8.9&	14.3&	11.6&	6.4&	10.0&	14.3&	6.9&	8.7&	11.3	\\ 
\multicolumn{1}{c|}{Qwen-VL-Max \cite{bai2023qwenvl}} &15.6&	10.7&	19.1&	20.0&	16.9&	12.5&	\textbf{17.9}&	16.4&	12.2&	\textbf{21.0}&	13.3&	14.2&	19.8&	11.5&	20.7&	13.0&	17.3	\\ 
\multicolumn{1}{c|}{Gemini Pro \cite{team2023gemini}} & 17.7&	15.1&	10.7&	20.7&	20.1&	11.9&	7.5&	20.2&	\textbf{21.1}&	16.8&	19.1&	19.0&	20.0&	14.3&	13.8&	17.4&	20.8     \\

\multicolumn{1}{c|}{GPT-4V \cite{gpt4v}} & \textbf{22.8}&	\textbf{27.3}&	\textbf{32.1}&	\textbf{35.7}&	\textbf{21.1}&	\textbf{16.7}&	13.4&	\textbf{22.1}&	14.4&	16.8&	\textbf{22.0}&	\textbf{22.2}&	\textbf{20.9}&	\textbf{23.8}&	\textbf{24.1}&	\textbf{21.7}&	\textbf{25.6}    \\
 \hline

\multicolumn{18}{c}{\multirow{1}{*}{\textit{Open-Source Multimodal Large Langugae Models (MLLMs)}}} \\ \hline

\multicolumn{1}{c|}{SPHINX-V2-13B \cite{lin2023sphinx}} &9.7&	6.7&7.1&	12.9&	7.5&	7.7&	6.0&	9.6&	\textbf{16.7}&	10.1&	11.0&	11.8&	12.5&	8.2&	8.6&	8.7&	6.0\\

\multicolumn{1}{c|}{LLaVA-1.5-13B \cite{liu2024llava-1.5}} & 11.1&	7.0&	14.3&	14.3&	9.1&	6.6&	6.0&	13.5&	5.6&	13.5&	10.4&	12.6&	14.7&	\textbf{11.5}&	13.8&	13.0&	10.7\\


\multicolumn{1}{c|}{\textbf{Math-LLaVA}} & \textbf{15.7}&	\textbf{9.0}&	\textbf{20.2}&	\textbf{15.7}&	\textbf{18.2}&	\textbf{10.1}&	\textbf{10.5}&	\textbf{16.4}&	14.4&	\textbf{16.0}&	\textbf{20.2}&	\textbf{18.4}&	\textbf{17.6}&	9.4&	\textbf{24.1}&	\textbf{21.7}&	\textbf{17.9}   \\ \hline

\end{tabular}}
\caption{Performance Comparison on the Math-V benchmark with the accuracy metric across various mathmatical subjects. Baseline results are obtained from \citet{wang2024mathv}.
The best results in both the close-source and open-source MLLMs are in bold.}
\vspace{-10pt}
\label{tab:math-v}
\end{table*}

\subsection{Generalizability of Math-LLaVA}
The proposed Math-LLaVA model has demonstrated exceptional performance in multimodal mathematical reasoning tasks. To assess its generalization capability, we conduct evaluation experiments using the MMMU benchmark, which encompasses various disciplines and domains. The results are shown in Table~\ref{tab:mmmubaseline}. With only the selected data, Math-LLaVA has a performance drop on science subset. However, we can observe that the Math-LLaVA model fine-tuned on MathV360K can significantly outperforms the base model, LLaVA-1.5-13B, as well as several other open-source MLLMs on all six sub-domains. This superior performance underscores its capability to generalize to downstream multimodal understanding and reasoning tasks. Furthermore, the fine-tuning process using our synthetic data does not detract from the model's reasoning abilities in other domains; rather, it enhances its generalizability.

\begin{table*}[h]\small
\centering

\renewcommand{\arraystretch}{1.2}
\setlength{\tabcolsep}{1.2mm}{
\begin{tabular}{c|c|cccccc}
\hline
				
			
\multicolumn{1}{c|}{\multirow{1}{*}{\textbf{Model}}} & \textbf{MMMU} & \makecell[c]{\textbf{Art \&}\\ \textbf{Design}} & \textbf{Business} & \textbf{Sci.}& \makecell[c]{\textbf{Health \&} \\ \textbf{Med.}} & \makecell[c]{\textbf{Human. \&} \\ \textbf{Social Sci.}} & \makecell[c]{\textbf{Tech. \&}\\ \textbf{Eng.}}\\ \hline 

\multicolumn{1}{c|}{Random Chance} & 22.1&	29.2&	24.7&	18.0&	20.7&	20.0&	21.4 \\
\multicolumn{1}{c|}{Frequent Guess} & 26.8 &	23.3&	\textbf{29.3}&	27.3&	30.0&	25.8&	24.8 \\
\multicolumn{1}{c|}{miniGPT4-7B} & 26.8 &	29.2&	21.3&	28.7&	30.7&	29.2&	23.8    \\
\multicolumn{1}{c|}{mPLUG-Owl-7B} & 32.7 &	45.8&	24.7&	22.7&	32.0&	45.8&	31.0    \\
\multicolumn{1}{c|}{SPHINX-13B} & 32.9&	48.3&	24.7&	26.7&	30.7&	50.0&	26.2    \\
\multicolumn{1}{c|}{InstructBLIP-7B} & 32.9& 	40.0& 	28.0& 	\textbf{32.7}& 	28.7& 	47.5& 	27.1    \\
\multicolumn{1}{c|}{LLaVA-1.5-13B} & 36.4&	51.7&	22.7&	29.3&	38.7&	53.3&	31.4    \\ \hline
\multicolumn{1}{c|}{Math-LLaVA-DS} & 36.9 &	\textbf{55.0} & 24.7	& 23.3 & 38.7	& 56.7	& 32.4 \\
\multicolumn{1}{c|}{Math-LLaVA} & \textbf{38.3} &	53.3 & 24.7	&30.7	 & \textbf{38.7}	& \textbf{58.3}	& \textbf{33.3}	  \\ \hline
\end{tabular}}
\caption{Comparison with baselines on the MMMU benchmark.}
\label{tab:mmmubaseline}
\end{table*}
\subsection{Overfitting to Text Modality}
The proposed data synthesis pipeline generates additional question-answer pairs for each image to enhance the mathematical reasoning of MLLMs. Intuitively, we should investigate whether the proposed model, Math-LLaVA, is overfitting on the generated question-answer pairs. If overfitting occurs, Math-LLaVA might memorize or retrieve image information without requiring any visual input. To examine this, we compare the performance of Math-LLaVA before and after data synthesis, referred to as Math-LLaVA-DS and Math-LLaVA, respectively, on MathVista using text inputs only. As shown in Table \ref{tab:text overfit}, Math-LLaVA exhibits similar performance, approximately 32.0\%, as Math-LLaVA-DS on MathVista when inference is performed without any visual information. Furthermore, fine-tuning Math-LLaVA with only text data also yields similar observations. This indicates that the Math-LLaVA model is not overfitting on the synthesized question-answer pairs.

Interestingly, we also observe that with text-only input, LLaVA-1.5-13B achieves an accuracy of 23.3\% on MathVista. 
Potential reasons as explored in \cite{chen2024we} could be that visual content is unnecessary for many samples in MathVista and that unintentional data leakage may occur during the pre-training of LLMs and MLLMs.

\begin{table}[]\small
\centering
\renewcommand{\arraystretch}{1.3}
\setlength{\tabcolsep}{1.2mm}{
\begin{tabular}{c|c|c|c}
\hline

\multicolumn{1}{c|}{\multirow{1}{*}{\textbf{Model}}} &\textbf{Training}&\textbf{Inference} &\multicolumn{1}{c}{\textbf{MathVista}}   \\ \hline 
\multicolumn{1}{c|}{LLaVA-1.5-13B} & Image-Text & Text & 23.3   \\
\multicolumn{1}{c|}{Math-LLaVA-DS} & Image-Text & Text & 32.2   \\
\multicolumn{1}{c|}{Math-LLaVA} & Image-Text  & Text & 32.4 \\ \hline
\multicolumn{1}{c|}{Math-LLaVA-DS} & Text & Text & 32.1   \\
\multicolumn{1}{c|}{Math-LLaVA} & Text & Text & 32.5 \\

\hline
\end{tabular}}
\caption{Results of inference using only text of MathVista as input. Fine-tuning LLaVA-1.5 using image-text or text-only data.}
\vspace{-5pt}
\label{tab:text overfit}
\end{table}

\subsection{Effectiveness of Synthesis}
To verify the effectiveness of data selection and the proposed data augmentation strategies, we conduct experiments on various components of MathV360K independently. Initially, we fine-tune the LLaVA-1.5 model on 40K randomly sampled data points from the source dataset, without any selection, to demonstrate the efficacy of data filtering and proportioning. Subsequently, we separately combine the selected 40K data points with the generated data using four augmentation methods: mining images for QA generation (AskImg), posing complex questions (CompQ), rephrasing questions for logical consistency (RephQ), and simplifying questions for underspecification (SimpQ). Table~\ref{tab:each_effect} presents the accuracy achieved by different combinations of augmentations on MathVista. The results indicate that our data synthesis approach, which incorporates data selection and each augmentation method, yields better performance. Collectively, these strategies result in a significant 11\% improvement over randomly sampling 40K data points.

\begin{table}[h]\small
\centering

\renewcommand{\arraystretch}{1.2}
\setlength{\tabcolsep}{1.0mm}{
\begin{tabular}{ccccc|c}
\hline

\multicolumn{1}{c}{\textbf{Select}} & \multicolumn{1}{c}{\textbf{AskImg}} & \multicolumn{1}{c}{\textbf{CompQ}} &\multicolumn{1}{c}{\textbf{RephQ}} & \multicolumn{1}{c}{\textbf{SimpQ}} &\multicolumn{1}{|c}{\textbf{ALL}}\\ \hline 

\multicolumn{1}{c}{\XSolidBrush} &  {\XSolidBrush}& {\XSolidBrush} &{\XSolidBrush} &{\XSolidBrush} & 35.6   \\
\multicolumn{1}{c}{\CheckmarkBold} &  {\XSolidBrush}& {\XSolidBrush} &{\XSolidBrush} &{\XSolidBrush} & 38.2\\
\multicolumn{1}{c}{\CheckmarkBold} &  {\CheckmarkBold}& {\XSolidBrush} &{\XSolidBrush} &{\XSolidBrush} & 42.2\\
\multicolumn{1}{c}{\CheckmarkBold} &  {\XSolidBrush}& {\CheckmarkBold} &{\XSolidBrush} &{\XSolidBrush} & 39.8 \\
\multicolumn{1}{c}{\CheckmarkBold} &  {\XSolidBrush}& {\XSolidBrush} &{\CheckmarkBold} &{\XSolidBrush} & 40.9\\ 
\multicolumn{1}{c}{\CheckmarkBold} &  {\XSolidBrush}& {\XSolidBrush} &{\XSolidBrush} &{\CheckmarkBold} &41.1 \\
\multicolumn{1}{c}{\CheckmarkBold} &  {\CheckmarkBold}& {\CheckmarkBold} &{\CheckmarkBold} &{\CheckmarkBold} & \textbf{46.6}\\
\hline
\end{tabular}}
\caption{Effectiveness of data selection and different data augmentation strategies on MathVista.}
\vspace{-5pt}
\label{tab:each_effect}
\end{table}

\subsection{Enhancements from Augmentation of Each Task Type}
\begin{figure}[]
    \centering
   \includegraphics[width=1.0\columnwidth]{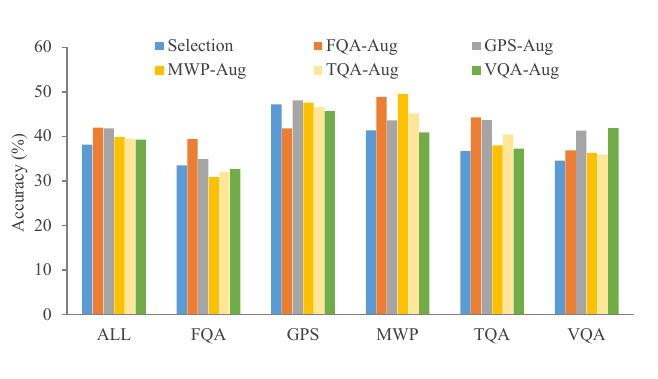}
    \caption{Accuracy on MathVista by augmentation for each task type.
    }
    \vspace{-10pt}
    \label{fig:effect-type}
\end{figure}

Given that we selected data from five different question-answering task types, our aim is to investigate which types or skills in multimodal mathematical reasoning could be enhanced by augmenting the source data from each individual task category. To this end, we conduct experiments with newly synthesized data for each task type, mixed with selected data. The results on MathVista are presented in Figure~\ref{fig:effect-type}. We observe that augmentation of various types of source data can further improve the model's performance on the corresponding tasks. The enhancements are particularly pronounced for tasks involving FQA, MWP, and VQA. Interestingly, data augmentation for a single task type also shows improvements in effectiveness for other task types, likely due to the overlap in reasoning skills required across different tasks.

\section{Conclusions}
We addressed the shortage of high-quality and diverse multimodal mathematical training datasets by creating MathV360K, which consists of 40K high-quality multimodal questions and answers from 24 existing datasets, along with 320K newly synthesized question-answer pairs. This comprehensive dataset enhances both the breadth and depth of multimodal mathematical questions. Using MathV360K, we fine-tuned Math-LLaVA, significantly improving its capability in multimodal mathematical reasoning, outperforming LLaVA-1.5 by 19 points on the minitest split of MathVista,
and yielding leading performance on Math-V and MathVerse.
Additionally, Math-LLaVA was validated on the MMMU benchmark, demonstrating its generalizability. Our research underscores the importance of dataset diversity and synthesis in enhancing the mathematical reasoning abilities of MLLMs.


\section{Limitations}
The data we selected and synthesized are in the format of images, questions, and answers, lacking intermediate steps that could be further improved. In future work, we will introduce annotated intermediate steps and rationale to construct more comprehensive and high-quality datasets to further enhance the MLLMs's multimodal reasoning capability.

\section{Ethics Statement}
We do not envision that our work will result in any harm as defined in ethics policy. LLaVA-1.5 base model uses LLaMA 2 Community License and ViT-Large-Patch16-224 uses Apache License 2.0. For datasets, GEOS, A-OKVQA and MMMU use Apache License 2.0. Geometry3K, FigureQA and PMC-VQA use MIT License. Super-CLEVR uses BSD License. ChartQA uses GPL 3.0 License. GeoQA+, UniGeo and DocVQA are publicly available for research purposes. The rest of the dataset use permissive Creative Commons Licenses. The intended use of these source datasets and evaluation datasets is to train and test the model's multimodal reasoning capability, which is consistent with our utilization of all these data. Our proposed MathV360K can improve the multimodal mathematical reasoning ability of the open-source LLaVA-1.5 through training. Our data and model are publicly available.

\section{Acknowledgement}
This work is supported by the National Natural Science Foundation of China under grant 62220106008, U20B2063, and 62102070. This research is supported by A*STAR, CISCO Systems (USA) Pte. Ltd and National University of Singapore under its Cisco-NUS Accelerated Digital Economy Corporate Laboratory (Award I21001E0002).
\bibliography{custom}

\appendix
\section{Appendix}
\subsection{Source Data Statistics}
We collected 24 visual question answering and multimodal mathematical reasoning datasets, each targeting a specific task type and visual content. We focused on five problem task types to compile the source dataset: Figure Question Answering (FQA), which involves analyzing charts and plots statistically; Geometry Problem Solving (GPS), which involves solving geometrical problems with diagrams and figures; Math Word Problem (MWP), which involves arithmetic calculations within the context of images; Textbook Question Answering (TQA), where reasoning is based on scientific knowledge and figures; and Visual Question Answering (VQA), which requires reasoning about objects, scenes, or relationships within images. These datasets from different domains can be combined to cover multiple tasks, incorporating diverse visual contexts and mathematical skills. Although TQA and VQA primarily involve questions about scenes and relationships, they also include questions requiring arithmetic or numeric skills. Such data enhances multimodal mathematical reasoning and generalizes to other question answering tasks.

The source data are summarized in Table~\ref{tab:sourcedata} corresponding to Section~\ref{Data Selection}.
\begin{table*}[t]\small
\centering
\renewcommand{\arraystretch}{1.4}
\setlength{\tabcolsep}{1.0mm}{
\begin{tabular}{c|c|c|c|c|cccc}
\hline

\multicolumn{1}{c|}{\multirow{2}{*}{\textbf{Dataset}}} & \multicolumn{1}{c|}{\multirow{2}{*}{\textbf{Task}}} &
\multicolumn{1}{c|}{\multirow{2}{*}{\textbf{Visual Context}}} &
\multicolumn{1}{c|}{\multirow{2}{*}{\textbf{Training Images}}}&
\multicolumn{1}{c|}{\multirow{2}{*}{\textbf{Clear Images}}}& \multicolumn{4}{c}{\multirow{1}{*}{\textbf{Image Complexity}}}\\ \cline{6-9}
\multicolumn{1}{c|}{} & \multicolumn{1}{c|}{} &\multicolumn{1}{c|}{} &\multicolumn{1}{c|}{} &\multicolumn{1}{c|}{} & 0& 1& 2& 3         \\ \hline


DocVQA \citeyearpar{mathew2022infographicvqa} &	FQA & Document Image&	8535 &8227	&2086&	6007&	125&	9	\\
FigureQA \citeyearpar{kahou2017figureqa}&	FQA&	Charts and Plots&	18173	&	18173&	687	&16792&	694&	0\\
DVQA \citeyearpar{kafle2018dvqa}&	FQA&	Bar Chart	&	19092&	19092&	21&	18021&	1045&	5\\
PlotQA \citeyearpar{methani2020plotqa}&	FQA&	Bar, Line, Scatter & 18782	&	18782&	13&	18759&	10&	0\\
ChartQA \citeyearpar{masry-etal-2022-chartqa}	&	FQA&	Charts and Plots&	3699& 3699&		0&	3649&	50&	0\\
MapQA \citeyearpar{chang2022mapqa}&FQA	&Map Chart&	10020 &	10016&	1&	10015&	0&	0\\ \hline

IconQA \citeyearpar{lu2021iconqa}&	MWP&	Abstract Scene&	20000&	19068&	10991&	8055&	22&	0\\
CLEVR-Math \citeyearpar{lindstrom2022clevr}&	MWP&	Synthetic Scene&	17552&	17551&	1&	17550&	0&	0\\
TabMWP \citeyearpar{lu2022dynamic}&	MWP	&Table &	20000 &20000&	14919&	5081&	0&	0 \\ \hline
GEOS \citeyearpar{seo-etal-2015-solving}	&	GPS	&Geometry Diagram	&66	&64	&2&	57&	5&	0 \\
Geometry3K \citeyearpar{lu-etal-2021-inter} &	GPS&	Geometry Diagram&	2101 &	2101&	21	&1508&	568&4 \\
GeoQA+ \citeyearpar{cao-xiao-2022-augmented}& GPS&	Geometry Diagram&	6027&	5956&	103&4399&	1454&	0\\
UniGeo \citeyearpar{chen-etal-2022-unigeo} 	&	GPS	&Geometry Diagram&	3499&	3432&	72&	2514&	846	&0 \\ \hline
TQA	\citeyearpar{kembhavi2017you}& TQA&	Scientific Figure&	1499&		1497&	20&	949&	498&	30\\
AI2D \citeyearpar{kembhavi2016diagram}&TQA	&Scientific Figure&	3247&	3235&	32&	2321&	823&	59\\
ScienceQA \citeyearpar{lu2022scienceqa}	&TQA&	Scientific Figure&	6218&6061&	1533&	4251&	273&	4\\ \hline
A-OKVQA \citeyearpar{schwenk2022okvqa}	&VQA&	Natural Image	&	16540 &	14526&	10&	11724&	2743&	49\\
VQA2.0 \citeyearpar{goyal2017vqa20}	&	VQA&	Natural Image &	16912&	14521&	45&	12783&	1672&	21\\
PMC-VQA \citeyearpar{zhang2023pmc}	&	VQA	&Medical Image&	19682&	9846&	62&	2989&	3501&	3294\\
VizWiz \citeyearpar{gurari2018vizwiz}	&VQA&	Natural Image&	20,000&		16400&	790&	14800&	770&	40\\
Super-CLEVR	\citeyearpar{li2023super}&	VQA	&Synthetic Scene&		2000&	1950&	1&	1568&	381&	0\\
VQA-AS \citeyearpar{antol2015vqaas}	&VQA	&Abstract Scene	&	14065&		14065&	7&	13996& 62&	0 \\
VQA-RAD	\citeyearpar{lau2018vqarad} &	VQA&	Medical Image&		259&	248&	0&	91&	95&	62\\
TextVQA \citeyearpar{singh2019textvqa}	&VQA&	Natural Image&	15815&	11350&	179&	9497&	1598&	76\\

 \hline
\end{tabular}}
\caption{Summary of the 24 different source traing datasets for collection. The table provides details on their task, visual context, distribution of image clarity and comprehension complexity according to fine-tuned ViT classification model.
Among them, only the text data of GeoQA+ are in Chinese, the rest source data are in English.}
\label{tab:sourcedata}
\end{table*}

\subsection{Results on MathVerse Benchmark}
The proposed Math-LLaVA model has demonstrated impressive performance on MathVista and Math-V. To assess its multimodal mathematical reasoning capabilities more comprehensively, we conduct evaluation experiments using the MathVerse benchmark \cite{zhang2024mathverse}.
The results are shown in Table~\ref{tab:mathverse}.
Math-LLaVA also achieves significant improvement compared to the base model and impressive performance among most open-source MLLMs.

\begin{table*}[!h]\small
\centering

\renewcommand{\arraystretch}{1.4}
\setlength{\tabcolsep}{1.6mm}{
\begin{tabular}{c|c|c|c|c|c|c}
\hline

\multicolumn{1}{c|}{\multirow{2}{*}{\textbf{Model}}} & \multicolumn{6}{c}{\textbf{MathVerse}} \\ \cline{2-7}
\multicolumn{1}{c|}{} &ALL &\makecell[c]{Text \\Dominant} &	\makecell[c]{Text\\ Lite} &	\makecell[c]{Vision\\ Intensive}&	\makecell[c]{Vision\\ Dominant}&	\makecell[c]{Vision\\ Only}         \\ \hline

\multicolumn{7}{c}{\multirow{1}{*}{\textit{Heuristics Baselines}}} \\ \hline
\multicolumn{1}{c|}{Random Chance} &12.4 & 12.4& 12.4& 12.4& 12.4& 12.4    \\
\multicolumn{1}{c|}{Human} & 64.9	& 	71.2	& 	70.9& 	61.4	& 	68.3	& 	66.7         \\ \hline

\multicolumn{7}{c}{\multirow{1}{*}{\textit{Close-Source Multimodal Large Langugae Models (MLLMs)}}} \\ \hline

\multicolumn{1}{c|}{Qwen-VL-Plus \cite{bai2023qwenvl}} &11.8	&	15.7	&	11.1&			9.0&		13.0&		10.0	\\ 
\multicolumn{1}{c|}{Gemini Pro \cite{team2023gemini}} & 23.7&	27.6&	23.7&	19.4&	20.3&	20.5     \\

\multicolumn{1}{c|}{GPT-4V \cite{gpt4v}} & \textbf{38.9}&	\textbf{52.1}&	\textbf{40.9}&	\textbf{34.9}&	\textbf{33.6}& \textbf{29.8}    \\
 \hline

\multicolumn{7}{c}{\multirow{1}{*}{\textit{Open-Source Multimodal Large Langugae Models (MLLMs)}}} \\ \hline

\multicolumn{1}{c|}{mPLUG-Owl-7B \cite{ye2023mplug}} & 4.6&	6.6&	6.3&	6.3&	5.6&	4.9 \\
\multicolumn{1}{c|}{LLaMA-Adapter-V2-7B \cite{gao2023llama}} & 5.7&	6.2&	5.9&	6.1&	4.2&	6.1 \\

\multicolumn{1}{c|}{LLaVA-1.5-13B \cite{liu2024llava-1.5}} & 7.6&	8.8&	7.6&	7.4&	7.4&	6.9\\

\multicolumn{1}{c|}{SPHINX-V2-13B \cite{lin2023sphinx}} &12.2&	13.9&	11.6&	11.6&	13.5&	10.4\\


\multicolumn{1}{c|}{\textbf{Math-LLaVA}} & \textbf{19.8}&	\textbf{22.3}&	\textbf{21.6}&	\textbf{20.8}&	\textbf{19.2}& \textbf{15.2}   \\ \hline

\end{tabular}}
\caption{Performance Comparison on the MathVerse benchmark with the accuracy metric. Baseline results are obtained from \citet{zhang2024mathverse}.
The best results in both the close-source and open-source MLLMs are in bold.}
\label{tab:mathverse}
\end{table*}

\subsection{Distribution Proportioning of Image Comprehension Complexity}
We select images from the source data based on an overall complexity ratio of 2:3:4:1. Due to the limited number of the most complex images, all images with complexity level 3 are sampled. We employ a progressive distribution scale from easy to complex, as described in Section~\ref{Image Filtering and Proportioning}. In this section, we examine the impact of varying distribution proportions of the first three image comprehension complexity levels on model performance. We explore settings with different proportions of comprehension complexities 0, 1, and 2, including uniform distribution, decreasing proportions as complexity increases, and proportions that fluctuate with complexity. As demonstrated in Table~\ref{tab:Proportioning}, both uniform distribution of image complexity and decreasing proportion with increasing difficulty are less effective compared to a progressive proportional distribution aligned with complexity. These findings suggest that MLLMs require fewer simple images and question-answer pairs, but benefit from a larger proportion of complex training data to enhance multimodal mathematical reasoning.

\begin{table}[H]\small
\centering
\renewcommand{\arraystretch}{1.3}
\setlength{\tabcolsep}{1.2mm}{
\begin{tabular}{c|c|ccccc}
\hline

\multicolumn{1}{c|}{\multirow{1}{*}{\textbf{Proportion}}} &\multicolumn{1}{c|}{\textbf{ALL}} & \multicolumn{1}{c}{\textbf{FQA}} &\multicolumn{1}{c}{\textbf{GPS}} &\multicolumn{1}{c}{\textbf{MWP}} &\multicolumn{1}{c}{\textbf{TQA}} &\multicolumn{1}{c}{\textbf{VQA}}\\ \hline 
\multicolumn{1}{c|}{3:3:3:1} &  36.0 & 29.0 &44.4& 40.9&35.6  &34.5  \\
\multicolumn{1}{c|}{4:3:2:1} &  36.4 & 32.0 & 39.6&42.5&36.9 &35.1  \\
\multicolumn{1}{c|}{2:4:3:1} &  35.1 & 32.0 & 40.5 & 35.5 & 36.2 & 34.6   \\
\multicolumn{1}{c|}{2:3:4:1} &38.2 &33.5& 47.2& 41.4& 36.7& 34.6	   \\
\hline
\end{tabular}}
\caption{Comparison with different distribution proportioning of image comprehension complexity on MathVista.}
\label{tab:Proportioning}
\end{table}

\subsection{Cases Study} 
\begin{figure*}[t]
    \centering
    \includegraphics[width=1.0\textwidth]{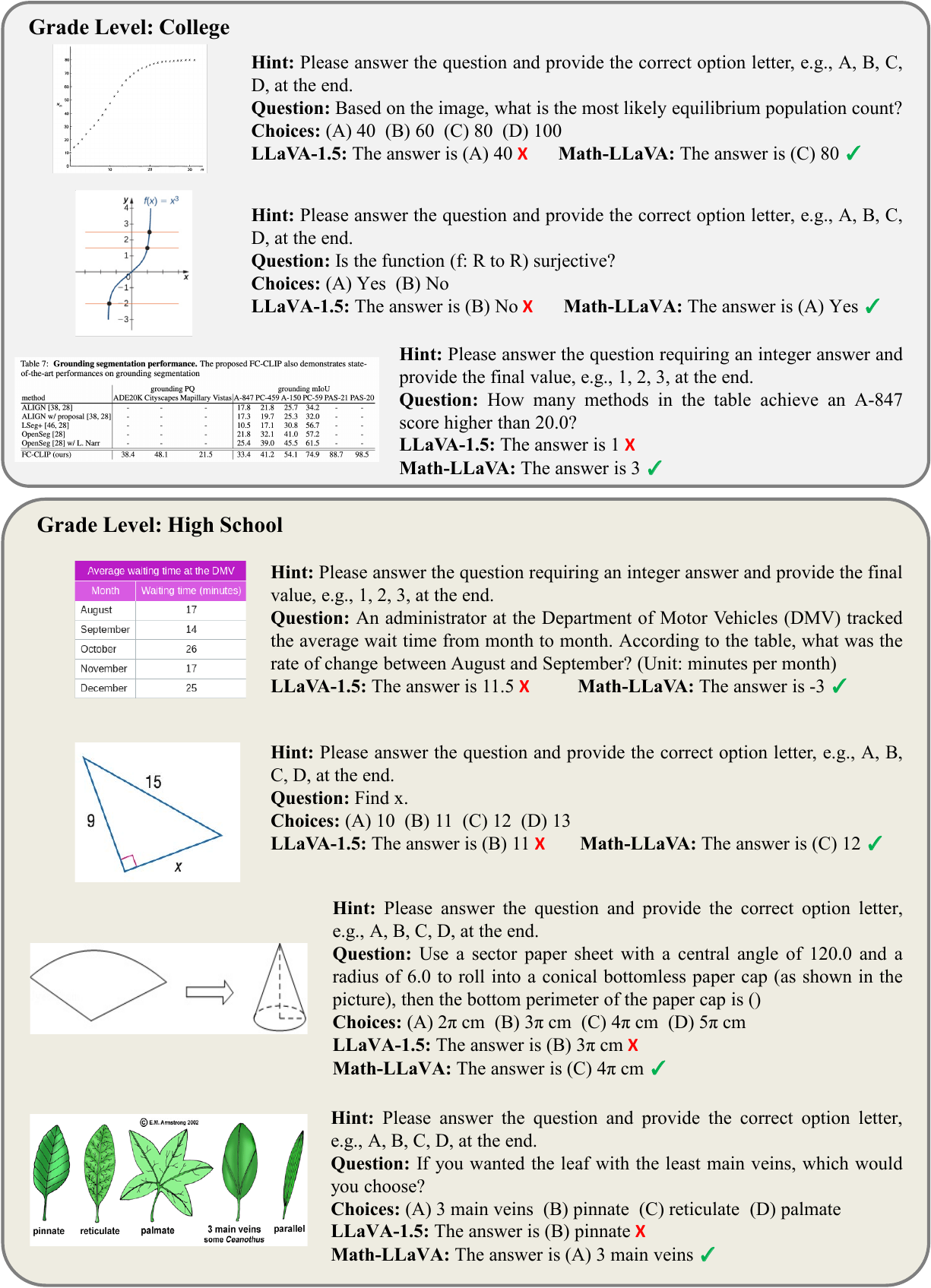}
    \caption{Cases on college and high school difficulty problems of MathVista, Math-LLaVA can solve better compared to LLaVA-1.5.
    }
    \label{fig:case_high}
\end{figure*}

We present several examples of solutions generated by Math-LLaVA and LLaVA-1.5 for image-question pairs of high school or college-level difficulty in MathVista. As illustrated in Figure~\ref{fig:case_high}, the base model (LLaVA-1.5) often performs inadequately on numerical computations involving tables, geometric problems, and counting tasks at the high school level. In contrast, our Math-LLaVA model demonstrates superior proficiency in addressing these high school problems, thanks to its training on selected and synthesized data designed to tackle complex issues. Although LLaVA-1.5 faces challenges when dealing with more advanced functions and detailed tables, Math-LLaVA shows promise and capability in solving such intricate problems.

\begin{figure*}[t]
    \centering
    \includegraphics[width=1.0\textwidth]{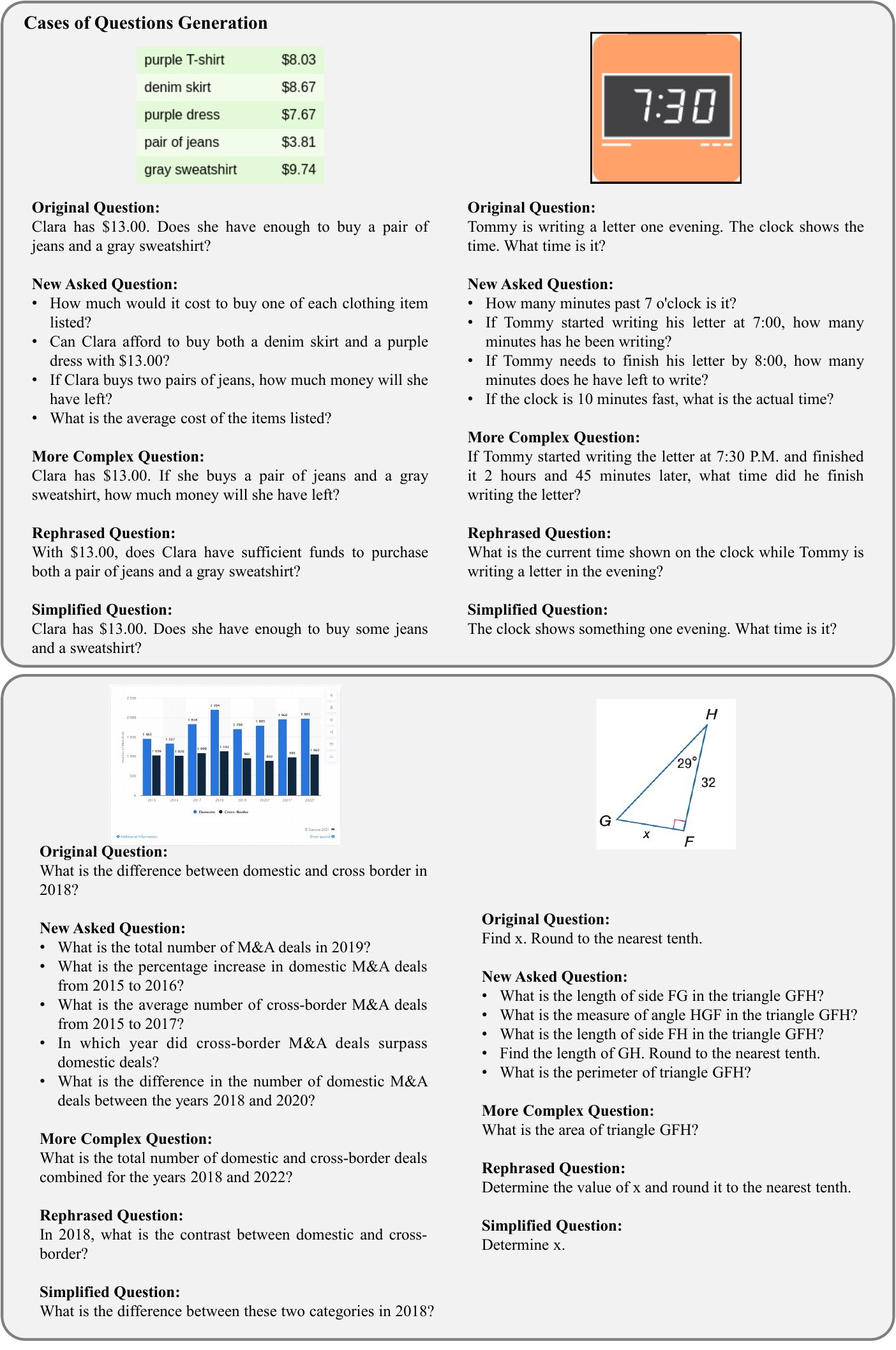}
    \caption{Examples of synthesizing new questions on source data.
    }
    \label{fig:case_new_question}
\end{figure*}

Additionally, we present several
examples of newly generated questions, created by thoroughly mining images and questions from the selected dataset. As depicted in Figure~\ref{fig:case_new_question}, existing dataset contains a limited number of image-question pairs. By fully utilizing the visual information from the images, we are able to generate a wider variety of questions from different perspectives, thereby enhancing the diversity of the problem set. The 
generated questions are created in a few-shot manner, referencing the format of existing question types. Consequently, these questions encompass more than just isolated visual content; they involve reasoning with the images. Moreover, the inclusion of complex questions, logically consistent rephrased questions, and simplified, underspecified questions increases the diversity and robustness of the dataset in terms of both question depth and format, compared to the original set of questions.

\begin{figure*}[t]
    \centering
    \includegraphics[height=1.52\textwidth, width=0.95\textwidth]{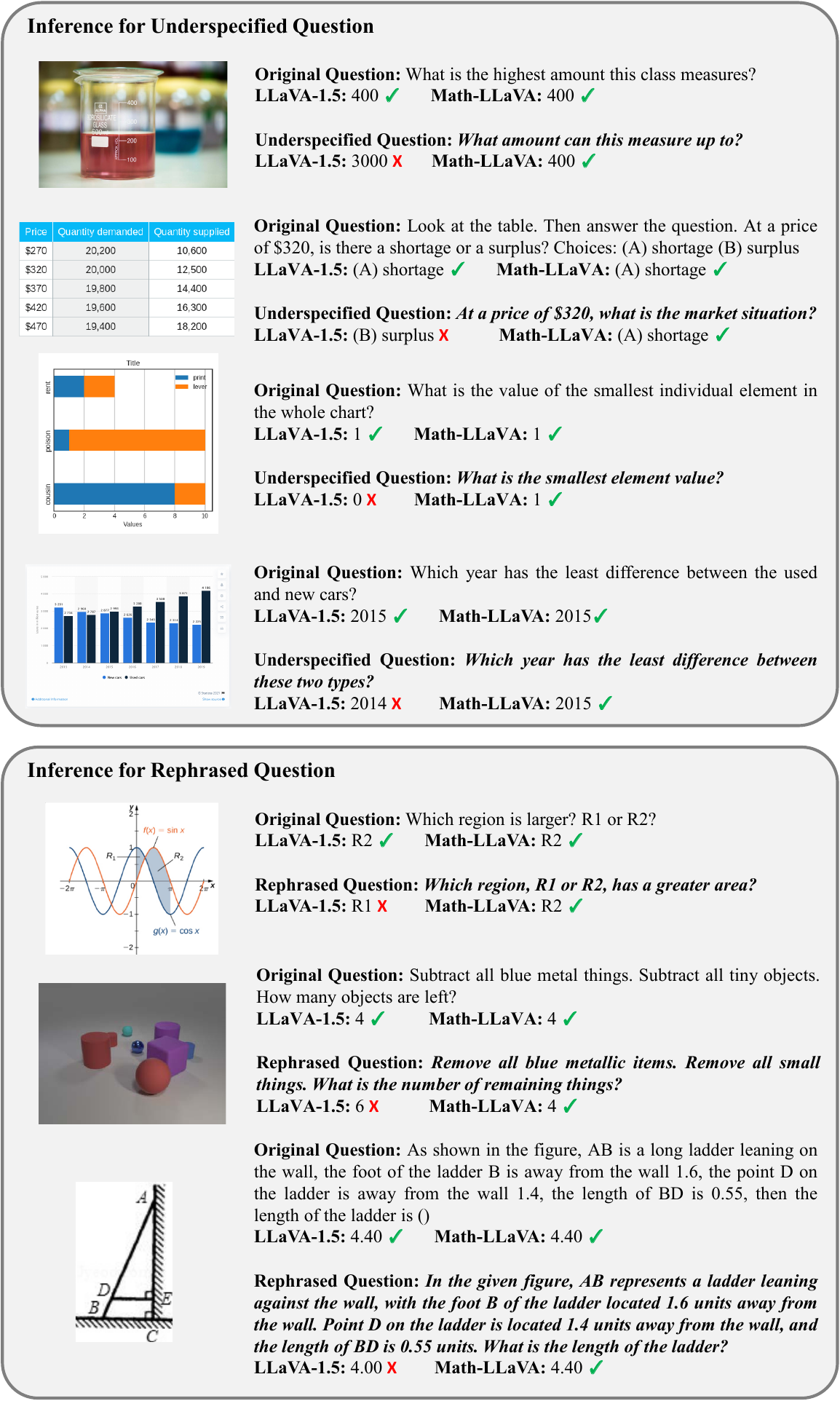} 
    \caption{Examples of testing on underspecified and rephrased questions.
    }
    \label{fig:case_under_reph}
\end{figure*}

Interestingly, Multimodal Language Models (MLLMs) demonstrate biases when handling multimodal mathematical reasoning tasks, particularly with logically consistent rephrased or underspecified questions. As illustrated at the top of Figure~\ref{fig:case_under_reph}, LLaVA-1.5 exhibits the ability to answer the original question correctly but tends to falter with simplified, underspecified questions. In contrast, Math-LLaVA proves to be more robust, consistently providing correct answers to underspecified questions. This trend is also observed with logically consistent rephrased questions. Therefore, the use of logically consistent and simplified underspecified questions for data augmentation can enhance the robustness of MLLMs in mathematical reasoning tasks.

\end{document}